\documentclass[a4paper]{article}

\usepackage{INTERSPEECH2022}
\usepackage[utf8]{inputenc}

\usepackage{listings,xcolor}
\usepackage{multirow}
\usepackage[normalem]{ulem}
\useunder{\uline}{\ul}{}
\usepackage{color}

\usepackage{arydshln}
\usepackage{graphicx}
\usepackage{todonotes}
\newcounter{todocounter}

\title{Seq-2-Seq based Refinement of ASR Output for Spoken Name Capture}
\name{Karan Singla, Shahab Jalalvand, Yeon-Jun Kim, Ryan Price, Daniel Pressel, Srinivas Bangalore}

\address{Interactions, LLC, New Providence, NJ, USA}
\email{\{ksingla,sjalalvand,ykim\}@interactions.com}

\begin{document}

\maketitle
\begin{abstract}
Person name capture from human speech is a difficult task in human-machine conversations. In this paper, we propose a novel approach to capture the person names from the caller utterances in response to the prompt "say and spell your first/last name". Inspired from work on spell correction, disfluency removal and text normalization, we propose a lightweight Seq-2-Seq system which generates a name spell from a varying user input. Our proposed method outperforms the strong baseline which is based on LM-driven rule-based approach.
\end{abstract}


\section{Introduction}

In order to authenticate a user and provide personalized services, most enterprise virtual agents (EVA, also known as Intelligent virtual agents (IVA)) rely on capturing the name of the user accurately. However, for a multinational enterprise with a customer base varied in nationality and accents, the challenge of extracting the names of a customer from their spoken utterance accurately is immensely challenging. With repeated reprompting in an attempt to capture the user's names, most virtual agents deliver a frustrating user experience. In addition to large size of vocabulary of names, the problem is compounded by homophonic names -- names that sound the same but have distinct orthography~\cite{cole1995new, yu2003improved, raghavan2005matching, bruguier2016learning}. While general purpose transcription of speech has seen significant advance in the past decade, high accuracy recognition and extraction of names remains a persistent issue. 

A conventional pipeline of speech recognition (ASR) --  typically using name lists encoded in grammars, and/or supplemented with statistical language models (SLMs) -- are followed by spoken language understanding (SLU) systems that extract the names from the user's response to a prompt from a virtual agent. In order to minimize the error in recognition and extraction of names, designers of speech interfaces to EVA often design prompts that request the user not only to say their first or last name but spell it as well. With the spelling of names, similar sounding names such as ``Stuard'', ``Stuart'', and ``Stewart'', might be correctly captured, as attested by the experimental results in this paper. Such techniques are employed in human-human conversations as well, to minimize errors in capture of names between interlocutors.

\begin{figure}[t]
\centering
\includegraphics[width=80mm,height=27mm]{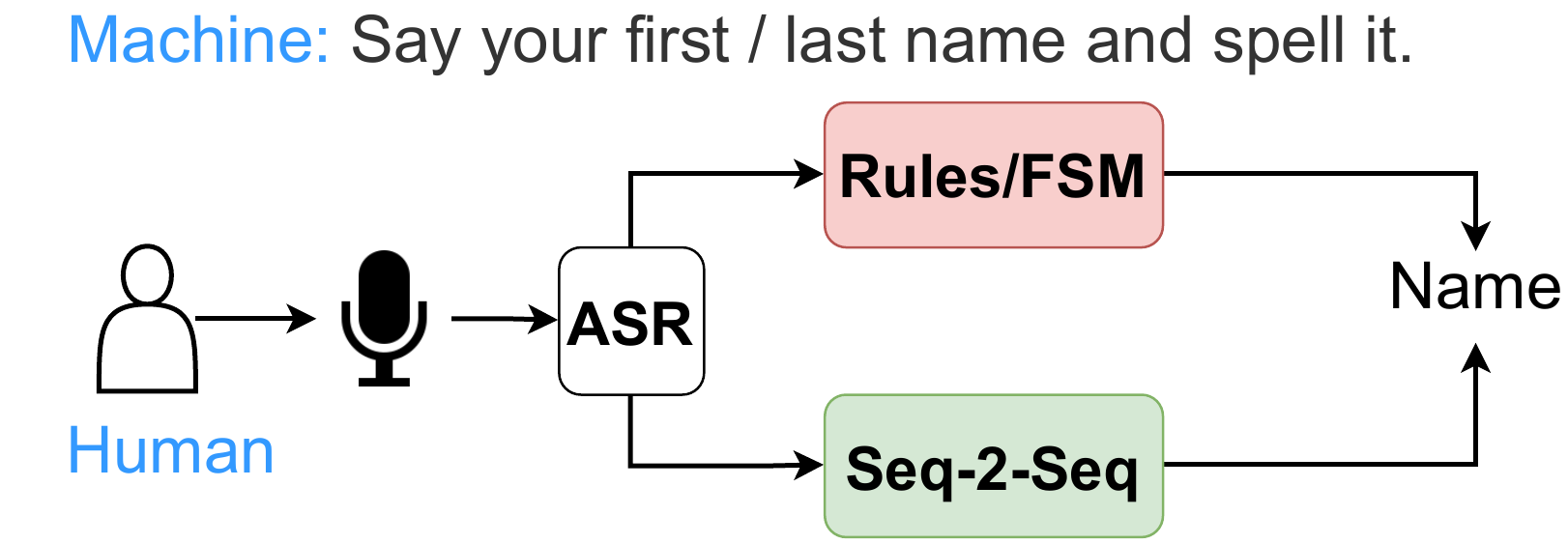}
\caption{Automatic name capture pipeline using sequence-2-sequence system instead of FSM or rule-based approaches.}
\end{figure}

\lstset{
    string=[s]{"}{"},
    stringstyle=\color{blue},
    comment=[l]{:},
    commentstyle=\color{black},
}


Spoken person name recognition combines ASR and NLP and traces its history to the early nineties with the first collection of spoken and spelled names over the telephone in 1992~\cite{cole1992telephone} in English and more recently, in languages such a Dutch~\cite{heuvel2008autonomata}. There have been several approaches~\cite{hild1996recognition} that leverage the spelled letters to alleviate the problem of open vocabulary nature of name recognition. \cite{jouvet1999recognition} leverage spelled letters to do alter the character sequence using a Finite State Machine (FSM) based approach. They use forward-backward algorithm to learn transition probabilities for editing spelled characters. Edited spelled name is then derived using A* search algorithm. They reject 10-15\% data with high conversational variation (including descriptors like \textit{as in, like}) as a part of incorrect data. 

Different designs have been proposed to capture names in dialogue. \cite{davidson2004usability} present usability evaluation of three different dialogue designs including say only, one stage say and spell and two-stage say and spell. \cite{wang2004chinese} use a combination of character description recognition and syllable spelling recognition to predict Chinese names. The most recent work on spoken say and spelled name recognition~\cite{price2021hybrid} propose use of dynamic hierarchical language model to accurately recognise the spelling and an LM-driven rule-based approach is used to predict the name. While these research works used hybrid ASR systems (a combination of neural network-based acoustic model with an n-gram language model), a recent approach~\cite{peyser2020improving} modified the loss criteria to specifically emphasize proper noun recognition in the context of an end-to-end ASR system. 




In recent work~\cite{guo2019spelling, hrinchuk2020correction}, Seq-2-Seq models have successfully been used for correction of ASR errors resulting from E2E ASR models. In such approaches, the first-pass hypothesis from ASR is given to an attention-based, encoder-decoder Seq-2-Seq model which outputs a version of the hypothesis, correcting any recognition errors, if present. Our proposed approach also applies Seq-2-Seq to the text hypothesis output by an E2E ASR model in the first pass. However, unlike previous works for general correction of ASR errors, our approach both \textit{corrects and extracts} spoken names from the ASR hypothesis. Concretely, our approach generates a character sequence of the name from the ASR hypothesis.

Our proposed Seq-2-Seq modeling for name capture is a light-weight adaptation of standard transformer based Seq-2-Seq architecture, similar to the one proposed by \cite{vaswani2017attention} for neural machine translation. This Seq-2-Seq system learns using ({\em asr hypothesis, name}) pairs recorded from an enterprise-grade virtual assistant. 
We believe the proposed approach can be generalized to capture other difficult entities such as emails, and addresses from user's speech to a virtual agent.

\section{Automatic Name Capture}

The goal of our work is to capture user's name from his/her spoken utterance that is in response to a virtual agent's prompt: \textit{Please say and spell your first/last name}. The table below shows the ASR hypothesis and intended names, illustrating the need for an automatic name capture system to not only correct the incorrect character sequence produced by ASR, but also to ignore or interpret the military alphabet and complex patterns associated with spelled letters.

\begin{table}[ht]
\centering
\begin{tabular}{l|l}
\textbf{ASR Hypothesis}                & \textbf{Name} \\ \hline
jennifer j e n n i s e r               & jennifer      \\
d a r e n darren                       & daren         \\
sara s a r a last name weber w e v e r & sara          \\
v as victor e r as robert azera        & vera         
\end{tabular}
\caption{Sample name capture data.}
\end{table}

Our system is a pipeline of ASR and an NLU module. The ASR recognizes the user's utterance to generate N-best hypotheses and the NLU module extracts the caller's name from the hypotheses. In this paper, we compare two ASR architectures and two different NLU approaches for name capture. In addition, we experiment with different confidence scoring methods to improve the precision of name extraction at the cost of rejecting the low confidence results. 

\subsection{Seq-2-Seq models to translate ASR Hypothesis to Names}

Our Seq-2-Seq system generates spelled name from  ASR hypothesis. We adopt a standard transformer based encoder-decoder setup to train from ({\em ASR hypothesis, Name}) pairs. The ASR hypothesis is provided in the form of BPE tokens as input to the Seq-2-Seq model, while the decoder generates a character sequence representing the name. We found that generating names as character sequence performs better than using BPEs for the output sequence. We use a shared embedding layer for both encoder and decoder tokens.

Our encoder-decoder setup is an adaptation  of \cite{vaswani2017attention} originally proposed for neural machine translation. The encoder is comparatively lightweight with a stack of N(=2) identical layers. Each layer first has a multi-head self-attention mechanism \cite{luong2015effective, shaw2018self} with 2 heads, and the second is a simple, position-wise, fully-connected, feed-forward network. Similar to \cite{vaswani2017attention}, we employ a residual connection \cite{he2016deep} around each of the two sub-layers. But instead of normalizing after each sub-layer, following \cite{nguyen2019transformers}, we perform PreNorm where we perform layer normalization before attention layers and before the fully connected dense layer. To facilitate residual connections, all sub-layers in the model, as well as the embedding layers, produce outputs with a hidden dimension of 64. We use fastBPE \cite{sennrich2015neural} to learn shared vocabulary for both encoder and decoder.

The decoder is also composed of a stack of N (=2) identical layers. In addition to the two sub-layers in each encoder layer, the decoder inserts a third sub-layer, which performs multi-head attention over the output of the encoder stack. Similar to the encoder, we use normalization employ residual connections around each of the sub-layers. We also modify the self-attention
sub-layer in the decoder stack to prevent positions from attending to subsequent positions. This
masking, combined with the fact that the output embeddings are offset by one position, ensures that the predictions for position $i$ can depend only on the known outputs at positions less than $i$. 
We use Adam with a fixed batch size of 32 and with a fixed learning rate of $1.0e-5$. We do not perform any pre-training instead train our system from scratch.

In the Seq-2-Seq based approach, the confidence-score is the log-probability assigned to the character name sequence. 

\subsection{Baseline: LM Driven Rule-based Approach}

For our baseline system, we simply concatenate the spelled letters and filter out other carrier phrases in the ASR hypothesis, in order to capture the name \cite{price2021hybrid}. The confidence score for this prediction is computed by averaging the ASR word-confidence scores of the spelled letters. 
If the predicted name matches any of the words in the utterance, then we consider it as a correct prediction, and set the confidence score of the prediction to 1.0. 
For example, if the ASR hypothesis is \textbf{"john/0.01 j/0.7 o/0.6 n/0.5 e/0.8"}, then the predicted label/score is \textbf{"jone/0.65"}. 

The overall process is as follows:
\begin{enumerate}
    \item Input the ASR hypothesis
    \item Find the spelled letters and concatenate them to predict the name
    \item Compute the confidence score by averaging the ASR scores of each spelled letters
    \item Set the confidence score to 1.0, if the predicted name matches a recognised word in the hypothesis;
    \item If there is no match using the 1-best hypothesis, then try the 2-best and 3-best hypotheses.
\end{enumerate}

However, there are some major drawbacks. The first drawback is its dependency on the letters. If the caller says \textbf{"tim t as in tango i as in i m as in man"}, then the recognized name will be \textbf{"tiim"}. Also the confidence scores become noisy as they also account for additional words which are not chars like \textbf{"tim, in, tango, as"}. Despite its simplicity, this approach is shown to be efficient and effective in our experiments, making it quite suitable for real-time EVA applications.

\section{Experiments and Results}
We collect name and spell data set from production IVA systems across a range of enterprise-grade virtual agents and measure the performances of different state-of-art techniques for name capture. We compare our proposed approach to a rule-based approach for automatic name capture.

\subsection{Dataset}

While the public datasets such as OGI collection~\cite{cole1995new} includes a small subset of spelled names, with the growing demand for EVAs for multiple industry verticals, we have millions of user utterances responding to the prompt {\em please say and spell your name} which have been annotated with the name spoken in the utterance.


We collect audio samples with corresponding name labels for both first name and last name capture tasks. In total, we collect 101K samples for first name and 580K samples for last names. Table \ref{tab:data} shows the statistics of the first and last name datasets, which illustrates that the variety of person names is quite large both in first names (26.2K) and last names (25.7K). Consequently, it is not feasible to use a simple classifier to predict the names \cite{price2021hybrid}. These datasets are collected from different applications including banking, insurance and retail from callers based in USA. We use 90\% of this data for training and 10\% for validation purposes. Number of words in train and development sets are average words in the ASR hypothesis from different ASR systems used in this paper and IA labels. 
For test purposes, we employ native speakers of English to annotate name labels. In total, there are 580 and 856 unique samples for first and last names respectively in the test set. We found there were around 10-15\% labeling errors corrected by annotators in the test set. 
For ASR model training purposes, we also collect additional 400 hours of human transcribed IVA data. 

\begin{table}[h]
    \centering
    \scalebox{0.93}{
    \begin{tabular}{l|c|c|c} 
    {\bf Dataset}  & {\bf \#Utterances} & {\bf \#Words} & {\bf \#UniqueNames} \\ \hline 
    IVA (Indomain) & 570K (800hrs) &  & -- \\ \hline
    firstname-train & 89K & 1.2M & 26.2K \\
    firstname-dev & 9.8K & 138.9K & 5K \\
    firstname-test  & 835   & 7.2K & 580 \\ \hdashline
    
    lastname-train  & 522K & 7.6M & 25.7K \\
    lastname-dev  & 58K & 882.5K & 12.6K \\
    lastname-test   & 1K   & 9.2K & 856   \\ 
    \end{tabular}
    }
    \caption{Data statistics.}
    \label{tab:data}

\end{table}



\subsection{ASR performance}

We compare two different architectures for transcribing spoken name capture utterances. First, we use a traditional hybrid DNN-HMM acoustic model with 4-gram language model and an FSM based decoder \cite{mohri2008speech,hinton2012deep}. Our second ASR system is a state-of-the-art end-2-end (E2E) deep residual convolutional neural network architecture \cite{majumdar2021citrinet}. It utilizes a Google sentence-piece \cite{kudo2018sentencepiece} tokenizer with vocabulary size of 1024, and transcribes text in lower case English alphabet along with spaces, apostrophes and a few other characters.
This model is pre-trained on ~7000 hours of publicly available transcribed data \cite{majumdar2021citrinet}. In our experiments, this model is fine-tuned on our in-domain IVA data to achieve the best performance on name capture. 

Based on ASR architectures and training corpora, we have the following acoustic models to evaluate.

\begin{itemize}
    \item {\bf Hybrid-indomain-AM} consists of hybrid DNN acoustic models trained to predict tied context-dependent triphone HMM states with cross-entropy and sequential loss functions using 81-dimensional log-spectrum features. This model is trained on  $\approx$ 400 hours of transcribed IVA speech data. We found that using in-domain dataset from IVA shows low error rate than using general purpose datasets available for ASR training.
    
    \item {\bf E2E-pretrained-AM} is the off-the-shelf Citrinet model \cite{majumdar2021citrinet} from Nemo toolkit \cite{kuchaiev2019nemo} with beam-search decoder (beam=64). No language model is used to rescore the hypotheses. 
    
    \item {\bf E2E-finetuned-AM} refers to the same Citrinet model, but fine-tuned on the  $\approx$ 400 hours of in-domain transcribed IVA dataset, same with \textit{Hybrid-indomain-AM} (refer Table 2).
    
\end{itemize}

We also experiment with two different language models:
\begin{itemize}

    \item \textbf{Indomain-LM} refers to a language model which is an interpolation of five 4-gram Katz backoff models \cite{katz1987estimation} each trained on a specific IVA vertical including banking, insurance, retail, hospitality and telecommunication. In total, we use a large text corpora which is obtained using an in-house production ASR containing 569 million words accounting for 41K unique words.
    \item \textbf{Names-LM} is a similar 4-gram Katz backoff model which uses additional data of name and spell capture described in Table \ref{tab:data}, along with Indomain IVA corpora. As shown in Table 2, using this LM shows relative improvement of 1-3\% in ASR Word-Error-Rate (WER) for name capture for both Hybrid and E2E ASR.

\end{itemize}

Fine-tuned E2E ASR shows significant error reductions from 27.6\% to 9.7\% for first name capture and 32.7\% to 7.9\% for last name capture. This highlights the need to fine-tune E2E ASRs on in-domain datasets with supervised transcription. Using \textit{name-LM} for rescoring the hypotheses significantly reduces the WER further, for the name capture test set by 0.5\%-1.3\%.

\begin{table}[ht]
\scalebox{0.9}{
\begin{tabular}{l|l|l|ll}
\textbf{}                                & \textbf{AM} & \textbf{LM} & \multicolumn{2}{l}{\textbf{\% Word Error Rate}} \\ \hline
          &          &       & \multicolumn{1}{l|}{{\ul FirstName}} & {\ul LastName} \\
\multicolumn{1}{c|}{\textbf{Hybrid ASR}} & Indomain                & Indomain                & \multicolumn{1}{l|}{{\ul}} 17.7   &  19.1    \\
          & Indomain & Names & \multicolumn{1}{l|}{}    14.7            &  19.1               \\ \hdashline
\multicolumn{1}{c|}{}    & Pretrained                 & No LM                   & \multicolumn{1}{l|}{}    27.6       &   32.7         \\
\textbf{E2E ASR} & Fine-tuned & No LM & \multicolumn{1}{l|}{}     9.7           &    7.9            \\
          & Fine-tuned & Names & \multicolumn{1}{l|}{}        {\bf 9.2}        &   {\bf 6.6}            
\end{tabular}
}
\caption{WER results on first and last name utterances.}
\label{tab:wer-results}
\end{table}

\begin{table*}[ht]
\centering
\begin{tabular}{l|l|l|ll|ll}
\textbf{} & {\bf AM} & {\bf LM} & \multicolumn{2}{l|}{{\bf FirstName}} & \multicolumn{2}{l}{{\bf LastName}} \\ \hline
                                                       &         &         & {\ul LM-driven} & {\ul Seq-2-Seq}        & {\ul LM-driven} & {\ul Seq-2-Seq}    \\
\multicolumn{1}{c|}{{\bf Hybrid ASR}}               & Indomain & Indomain &  37.8         & {\bf 29.2} &  31.0   &  {\bf 24.8}\\
                                                       & Indomain   & Names   & 30.9             & {\bf 24.4}       &  16.3          &    {\bf 13.1}    \\ \hdashline
\multicolumn{1}{c|}{\multirow{2}{*}{{\bf E2E ASR}}} & Pretrained & No LM   & 53.9             & {\bf 46.7}       &  62.5          &    {\bf 48.3}    \\
\multicolumn{1}{c|}{\multirow{2}{*}{}} & Fine-tuned & No LM   &  26.3            & {\bf 22.2}       & 12.2           &    {\bf 11.8}    \\
\multicolumn{1}{c|}{}                                  & Fine-tuned   & Names   & 25.3             & {\bf 21.9}       & {\bf 10.0}           &     11.0
\end{tabular}
\caption{1-best classification error rate scores.}
\label{tab:name_capture}
\end{table*}

In our experiments we attempted to fine-tune the AM with additional name and spell $\approx$ semi-supervised set. For this semi-supervised set, y samples, we choose the samples in which the concatenation of the spells match the name labels. We found that hybrid ASR performance improves for name and spell utterances but E2E ASR performance does not. Hence, we only use supervised indomain data for AM training. We also found that none of the above-mentioned ASR systems do a good job in recognising the name as a whole word because most of these names are not in-vocabulary for the language model nor are in the training set of the E2E ASR, consequently leading to errors shown in Table 3. Thus, callers are directed to spell their names, since spell recognition is more accurate, resulting in more reliable ways to predict the name \cite{levin2006learning}.

The WER results in Table~\ref{tab:wer-results} show that the end2end ASR outperforms over the traditional hybrid ASR when E2E model is fine-tuned with indomain data.
Even without using a language model, the fine-tuned end2end ASR provides 5-11.2\% absolute reduction in WER compared to the hybrid ASR.



\subsection{Results for name capture}

Table \ref{tab:name_capture} shows error rates for name capture on the first and last name test data. We achieve better results for name capture using the ASR system which had lower error rates, thus highlighting the importance of a high accuracy ASR system for name capture. Our proposed Seq-2-Seq approach performs significantly better than the LM-driven rule-based approach for the different ASR systems we compared. For dealing with hybrid ASR hypotheses, Seq-2-Seq captures 3-8\% names more correct than LM-driven approach.  Similarly name capture pipeline using E2E ASR shows 3-7\% relative improvement over the baseline. We notice that gains of Seq-2-Seq over LM-driven approach becomes larger, when ASR performance is poor.

LM-driven approach is highly dependent on ASR output, thus if the WER is high then name capture error will also be high. However, Seq-2-Seq seems to recover frequent ASR errors to capture names correctly. We believe seq-2-seq errors can be further reduced if the quality of training data is improved. Our study showed our training data has around 10-15\% labeling errors made by human annotators.

\section{Observations}

Table \ref{tab:exp} shows examples where seq-2-seq approach correctly filters input, corrects ASR errors and generates the correct letter sequence to capture names. It also shows examples where Seq-2-Seq performs worse than the baseline. We observe that Seq-2-Seq system for first name capture can filter-out first names from full name provided by speaker.

\definecolor{green}{rgb}{0.0, 0.5, 0.0}

\begin{table}[ht]
    \centering
    \scalebox{0.9}{
    \begin{tabular}{l|l|l}
    ASR hypothesis (Input)               & LM-driven & Seq-2-Seq  \\ \hline
    b as in boy o w d as in dog i c as in cat h & \textcolor{green}{bowdich} & \textcolor{green}{bowdich} \\
    jennifer j e n n i \uline{\textbf{s}} e r  & \textcolor{red}{jenniser} & \textcolor{green}{jennifer} \\
    r o s l i n d rislin r a n k i n franks & \textcolor{red}{roslindrankin} & \textcolor{green}{roslind} \\
    \uline{\textbf{r}}  \uline{\textbf{i}} rippe r i p p e e & \textcolor{red}{ririppee}  & \textcolor{green}{rippee}  \\
    sdov s e d o z & \textcolor{green}{sedoz} & \textcolor{red}{sedov} \\
   um um baskal b a s c a l & \textcolor{red}{bascal} & \textcolor{red}{bascal} \\

    \end{tabular}
    }
    \caption{Sample output for automatic name capture. Green means correct, red means name captured wrong.}
    \label{tab:exp}
\end{table}

In order to maximize the user experience of an EVA application, we have a unique human-in-the-loop solution which recruits humans just-in-time, when a model confidence is inadequate for automation. The confidence threshold determines the error versus rejection curve and a suitable operating point is chosen that optimizes the rejection at a given error rate.

Figure 2 and 3 shows the Error-Rejection (ER) curve for first  and last name test set. We sort all the samples with the confidence score for each name capture approach, then plot variation in error rates by removing samples with a given confidence score.

\begin{figure}[ht]
\vspace{-3.0mm}
\centering
  \includegraphics[clip,width=\columnwidth]{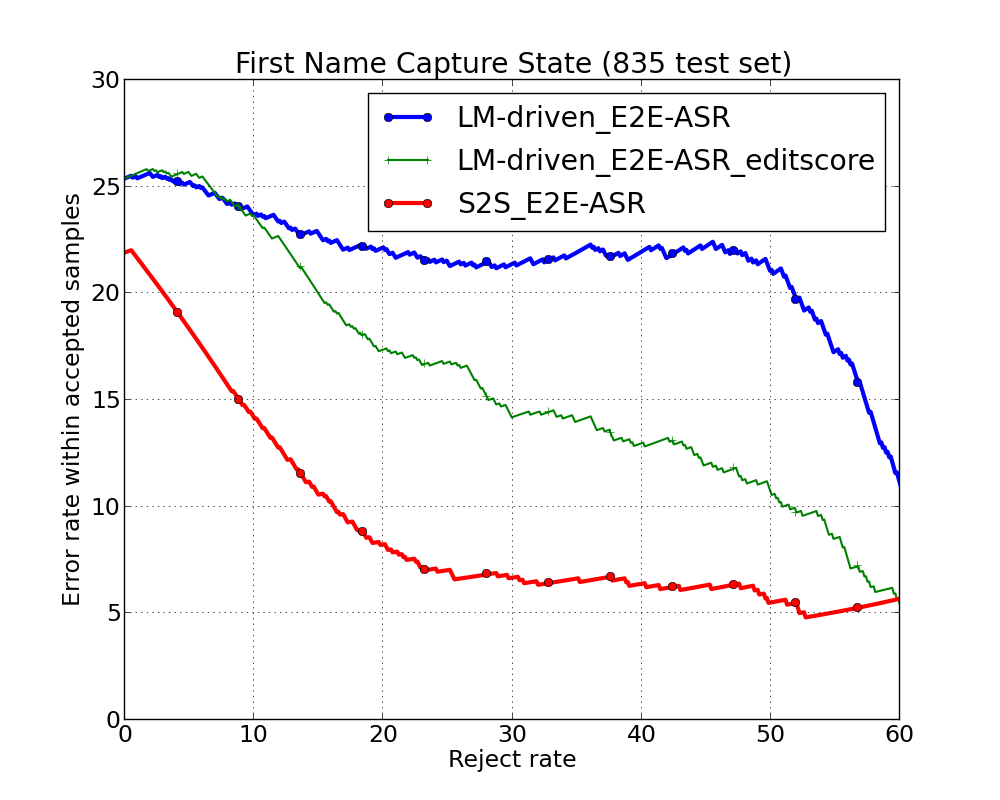}%
  \caption{Error vs Reject rate curve for First Name Capture.}
\end{figure}

\begin{figure}[ht]

  \includegraphics[clip,width=\columnwidth]{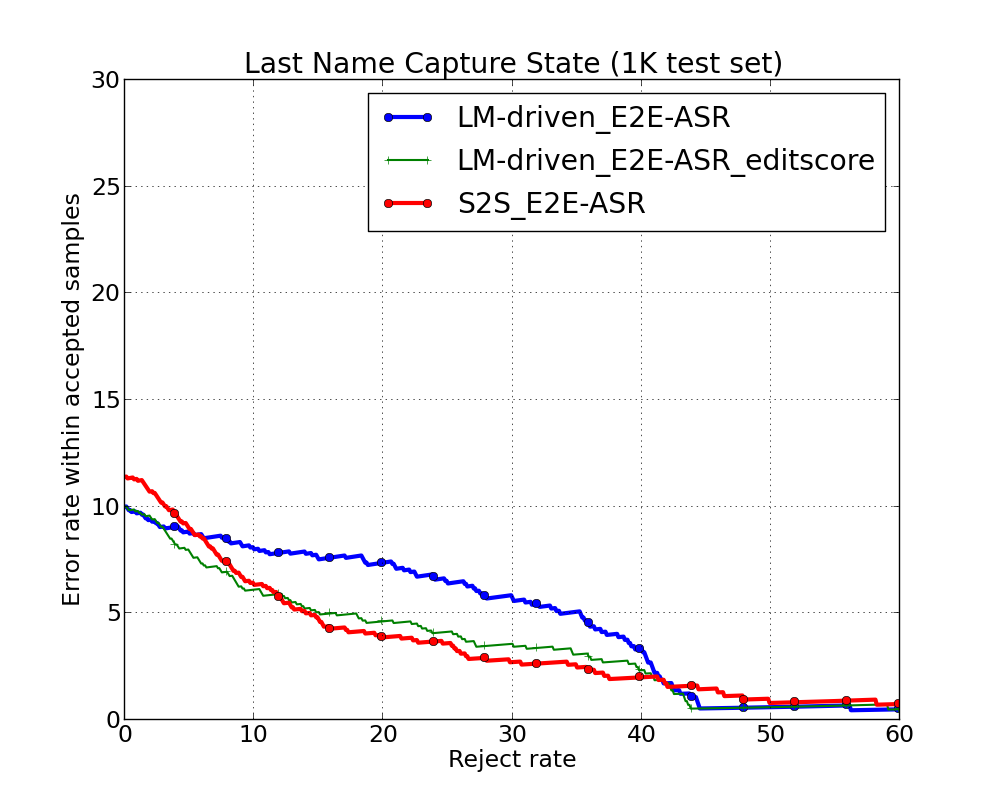}%
  \caption{Error vs Reject rate curve for Last Name Capture.}

\vspace{-4.0mm}
\end{figure}

The ER curve for first name capture shows Seq-2-Seq significantly improves over the baseline. For a low rejection rate of 20\% Seq-2-Seq performs at a significantly lower error rate of 8\% vs 21.5\% for the LM-driven approach. Similarly, for the last name capture, Seq-2-Seq shows significantly lower error rate at 20\% rejection rate. The big gap in performance could be due to the fact that confidence scores from LM-driven approach become noisy as they account for additional words which are not characters, as in \textbf{"tim, t, as, in, tango"}, unlike Seq-2-Seq which generates a letter sequence of caller's name. However when we replace LM-driven's confidence score with edit distance between the spelled name and first word in the utterance, LM-driven's performance improves (Blue vs. Green ER curves). This supports our earlier hypothesis that additional words in ASR hypothesis add noise to the confidence scores of LM-driven approach for name capture.

\section{Conclusions and Future Work}

Seq-2-Seq shows promising results to automatically capture first and last names from spoken utterances. We believe our results can have direct gains from exploiting ASR lattice output in form of N-best lists. We have not as yet extensively explored different hyperparamaters, thus, sweeping over certain model parameters can also lead to obvious improvements as well.

We believe the biases in user demographics can effect \textit{how we speak} \cite{stevens1992social}, therefore, we plan to do a wider and more thorough evaluation in the future. We will extend this approach to email and address capture prompts as we continue to explore Seq-2-Seq modeling techniques for data capture from spoken utterances.

\clearpage
\bibliographystyle{IEEEtran}
\bibliography{strings,refs}

\begin{thebibliography}{10}
\providecommand{\url}[1]{#1}
\csname url@samestyle\endcsname
\providecommand{\newblock}{\relax}
\providecommand{\bibinfo}[2]{#2}
\providecommand{\BIBentrySTDinterwordspacing}{\spaceskip=0pt\relax}
\providecommand{\BIBentryALTinterwordstretchfactor}{4}
\providecommand{\BIBentryALTinterwordspacing}{\spaceskip=\fontdimen2\font plus
\BIBentryALTinterwordstretchfactor\fontdimen3\font minus
  \fontdimen4\font\relax}
\providecommand{\BIBforeignlanguage}[2]{{%
\expandafter\ifx\csname l@#1\endcsname\relax
\typeout{** WARNING: IEEEtran.bst: No hyphenation pattern has been}%
\typeout{** loaded for the language `#1'. Using the pattern for}%
\typeout{** the default language instead.}%
\else
\language=\csname l@#1\endcsname
\fi
#2}}
\providecommand{\BIBdecl}{\relax}
\BIBdecl

\bibitem{cole1995new}
R.~A. Cole, M.~Noel, T.~Lander, and T.~Durham, ``New telephone speech corpora
  at cslu.'' in \emph{Eurospeech}.\hskip 1em plus 0.5em minus 0.4em\relax
  Citeseer, 1995, pp. 1--4.

\bibitem{yu2003improved}
D.~Yu, K.~Wang, M.~Mahajan, P.~Mau, and A.~Acero, ``Improved name recognition
  with user modeling,'' in \emph{Eighth European Conference on Speech
  Communication and Technology}, 2003.

\bibitem{raghavan2005matching}
\BIBentryALTinterwordspacing
H.~Raghavan and J.~Allan, ``Matching inconsistently spelled names in automatic
  speech recognizer output for information retrieval,'' in \emph{Proceedings of
  Human Language Technology Conference and Conference on Empirical Methods in
  Natural Language Processing}.\hskip 1em plus 0.5em minus 0.4em\relax
  Vancouver, British Columbia, Canada: Association for Computational
  Linguistics, Oct. 2005, pp. 451--458. [Online]. Available:
  \url{https://www.aclweb.org/anthology/H05-1057}
\BIBentrySTDinterwordspacing

\bibitem{bruguier2016learning}
T.~Bruguier, F.~Peng, and F.~Beaufays, ``Learning personalized pronunciations
  for contact names recognition,'' in \emph{Interspeech}, 2016.

\bibitem{cole1992telephone}
R.~A. Cole, K.~Roginski, and M.~A. Fanty, ``A telephone speech database of
  spelled and spoken names.'' in \emph{ICSLP}, vol.~92, 1992, pp. 891--895.

\bibitem{heuvel2008autonomata}
H.~Heuvel, J.-P. Martens, K.~D'hanens, and N.~Konings, ``The autonomata spoken
  names corpus,'' 2008.

\bibitem{hild1996recognition}
H.~Hild and A.~Waibel, ``Recognition of spelled names over the telephone,'' in
  \emph{Proceeding of Fourth International Conference on Spoken Language
  Processing. ICSLP'96}, vol.~1.\hskip 1em plus 0.5em minus 0.4em\relax IEEE,
  1996, pp. 346--349.

\bibitem{jouvet1999recognition}
D.~Jouvet and J.~Monn{\'e}, ``Recognition of spelled names over the telephone
  and rejection of data out of the spelling lexicon,'' in \emph{Sixth European
  Conference on Speech Communication and Technology}, 1999.

\bibitem{davidson2004usability}
N.~Davidson, F.~McInnes, and M.~A. Jack, ``Usability of dialogue design
  strategies for automated surname capture,'' \emph{Speech Communication},
  vol.~43, no.~1, pp. 55--70, 2004.

\bibitem{wang2004chinese}
N.~Wang, C.-H. Tsai, P.~Huang, and J.-L. Shen, ``Chinese large-vocabulary name
  recognition system using character description and syllable spelling
  recognition,'' in \emph{2004 International Symposium on Chinese Spoken
  Language Processing}, 2004, pp. 17--20.

\bibitem{price2021hybrid}
\BIBentryALTinterwordspacing
R.~Price, M.~Mehrabani, N.~Gupta, Y.-J. Kim, S.~Jalalvand, M.~Chen, Y.~Zhao,
  and S.~Bangalore, ``A hybrid approach to scalable and robust spoken language
  understanding in enterprise virtual agents,'' in \emph{Proceedings of the
  2021 Conference of the North American Chapter of the Association for
  Computational Linguistics: Human Language Technologies: Industry
  Papers}.\hskip 1em plus 0.5em minus 0.4em\relax Online: Association for
  Computational Linguistics, Jun. 2021, pp. 63--71. [Online]. Available:
  \url{https://aclanthology.org/2021.naacl-industry.9}
\BIBentrySTDinterwordspacing

\bibitem{peyser2020improving}
C.~Peyser, T.~N. Sainath, and G.~Pundak, ``Improving proper noun recognition in
  end-to-end asr by customization of the mwer loss criterion,'' in \emph{ICASSP
  2020-2020 IEEE International Conference on Acoustics, Speech and Signal
  Processing (ICASSP)}.\hskip 1em plus 0.5em minus 0.4em\relax IEEE, 2020, pp.
  7789--7793.

\bibitem{guo2019spelling}
J.~Guo, T.~N. Sainath, and R.~J. Weiss, ``A spelling correction model for
  end-to-end speech recognition,'' in \emph{ICASSP 2019-2019 IEEE International
  Conference on Acoustics, Speech and Signal Processing (ICASSP)}.\hskip 1em
  plus 0.5em minus 0.4em\relax IEEE, 2019, pp. 5651--5655.

\bibitem{hrinchuk2020correction}
O.~Hrinchuk, M.~Popova, and B.~Ginsburg, ``Correction of automatic speech
  recognition with transformer sequence-to-sequence model,'' in \emph{ICASSP
  2020-2020 IEEE International Conference on Acoustics, Speech and Signal
  Processing (ICASSP)}.\hskip 1em plus 0.5em minus 0.4em\relax IEEE, 2020, pp.
  7074--7078.

\bibitem{vaswani2017attention}
A.~Vaswani, N.~Shazeer, N.~Parmar, J.~Uszkoreit, L.~Jones, A.~N. Gomez,
  {\L}.~Kaiser, and I.~Polosukhin, ``Attention is all you need,''
  \emph{Advances in neural information processing systems}, vol.~30, 2017.

\bibitem{luong2015effective}
M.-T. Luong, H.~Pham, and C.~D. Manning, ``Effective approaches to
  attention-based neural machine translation,'' \emph{arXiv preprint
  arXiv:1508.04025}, 2015.

\bibitem{shaw2018self}
P.~Shaw, J.~Uszkoreit, and A.~Vaswani, ``Self-attention with relative position
  representations,'' \emph{arXiv preprint arXiv:1803.02155}, 2018.

\bibitem{he2016deep}
K.~He, X.~Zhang, S.~Ren, and J.~Sun, ``Deep residual learning for image
  recognition,'' in \emph{Proceedings of the IEEE conference on computer vision
  and pattern recognition}, 2016, pp. 770--778.

\bibitem{nguyen2019transformers}
T.~Q. Nguyen and J.~Salazar, ``Transformers without tears: Improving the
  normalization of self-attention,'' \emph{arXiv preprint arXiv:1910.05895},
  2019.

\bibitem{sennrich2015neural}
R.~Sennrich, B.~Haddow, and A.~Birch, ``Neural machine translation of rare
  words with subword units,'' \emph{arXiv preprint arXiv:1508.07909}, 2015.

\bibitem{mohri2008speech}
M.~Mohri, F.~Pereira, and M.~Riley, ``Speech recognition with weighted
  finite-state transducers,'' in \emph{Springer Handbook of Speech
  Processing}.\hskip 1em plus 0.5em minus 0.4em\relax Springer, 2008, pp.
  559--584.

\bibitem{hinton2012deep}
G.~Hinton, L.~Deng, D.~Yu, G.~E. Dahl, A.-r. Mohamed, N.~Jaitly, A.~Senior,
  V.~Vanhoucke, P.~Nguyen, T.~N. Sainath \emph{et~al.}, ``Deep neural networks
  for acoustic modeling in speech recognition: The shared views of four
  research groups,'' \emph{IEEE Signal processing magazine}, vol.~29, no.~6,
  pp. 82--97, 2012.

\bibitem{majumdar2021citrinet}
S.~Majumdar, J.~Balam, O.~Hrinchuk, V.~Lavrukhin, V.~Noroozi, and B.~Ginsburg,
  ``Citrinet: Closing the gap between non-autoregressive and autoregressive
  end-to-end models for automatic speech recognition,'' \emph{arXiv preprint
  arXiv:2104.01721}, 2021.

\bibitem{kudo2018sentencepiece}
T.~Kudo and J.~Richardson, ``Sentencepiece: A simple and language independent
  subword tokenizer and detokenizer for neural text processing,'' \emph{arXiv
  preprint arXiv:1808.06226}, 2018.

\bibitem{kuchaiev2019nemo}
O.~Kuchaiev, J.~Li, H.~Nguyen, O.~Hrinchuk, R.~Leary, B.~Ginsburg, S.~Kriman,
  S.~Beliaev, V.~Lavrukhin, J.~Cook \emph{et~al.}, ``Nemo: a toolkit for
  building ai applications using neural modules,'' \emph{arXiv preprint
  arXiv:1909.09577}, 2019.

\bibitem{katz1987estimation}
S.~Katz, ``Estimation of probabilities from sparse data for the language model
  component of a speech recognizer,'' \emph{IEEE transactions on acoustics,
  speech, and signal processing}, vol.~35, no.~3, pp. 400--401, 1987.

\bibitem{levin2006learning}
I.~Levin, S.~Shatil-Carmon, and O.~Asif-Rave, ``Learning of letter names and
  sounds and their contribution to word recognition,'' \emph{Journal of
  Experimental Child Psychology}, vol.~93, no.~2, pp. 139--165, 2006.

\bibitem{stevens1992social}
G.~Stevens, ``The social and demographic context of language use in the united
  states,'' \emph{American Sociological Review}, pp. 171--185, 1992.

\end{thebibliography}

\end{document}